\documentclass[10pt,twocolumn]{article}

\usepackage[margin=0.72in]{geometry}
\usepackage[T1]{fontenc}
\usepackage{lmodern}
\usepackage{microtype}
\usepackage{amsmath,amssymb}
\usepackage{booktabs}
\usepackage{threeparttable}
\usepackage{graphicx}
\usepackage{tabularx}
\usepackage{array}
\usepackage{multirow}
\usepackage{enumitem}
\usepackage{xcolor}
\usepackage{listings}
\usepackage{tikz}
\usetikzlibrary{arrows.meta,positioning,fit,calc}
\usepackage{pgfplots}
\pgfplotsset{compat=1.18}
\usepackage[numbers,sort&compress]{natbib}
\usepackage[
    colorlinks=true,
    linkcolor=black,
    citecolor=blue,
    urlcolor=blue
]{hyperref}
\usepackage{doi}
\usepackage{url}

\definecolor{codegray}{gray}{0.96}
\newcommand{\qwen}{Qwen2.5-Coder-7B}
\newcommand{\deepseek}{DeepSeek-Coder-V2-Lite}
\newcommand{\ifrate}{\mathrm{IF}}
\newcommand{\crrate}{\mathrm{CR}}
\newcommand{\fliprate}{\mathrm{Flip}}

\title{\textbf{Robustness of LLM-Generated SystemVerilog Assertions to Semantics-Preserving RTL Transformations}}
\author{FNU Aditi}
\date{}

\begin{document}
\maketitle

\begin{abstract}
Large language models (LLMs) are increasingly being explored for automating SystemVerilog Assertion (SVA) generation, yet most evaluations report correctness on a single syntactic representation of an input. Such point accuracy does not reveal whether a model's correct output is stable when the same RTL behavior is written differently. This paper presents a controlled metamorphic evaluation of LLM-based SVA generation under semantics-preserving RTL transformations. Starting from the VERT dataset, we construct a quality-filtered conditional-control pool and a stratified 40-program evaluation set containing 295 assignment behaviors. We evaluate two open code models, \qwen{} and \deepseek{}, with an identical evaluation prompt and greedy decoding. Three transformations are studied: operand reordering, deterministic identifier renaming, and redundant parenthesization. Beyond baseline and transformed accuracy, we measure conditional robustness, invariance failure, and any-flip rate, with 10,000-sample clustered bootstrap intervals at the RTL-program level. Across all six model--transformation conditions, 9.7\%--27.0\% of behaviors that were correct on the original RTL become incorrect after a semantics-preserving transformation. Aggregate accuracy can therefore hide substantial instability: under identifier renaming, \deepseek{} improves from 53.9\% to 63.7\% accuracy while 19.5\% of its originally correct behaviors fail. Manual review of 30 sampled correct-to-wrong transitions identifies dropped path predicates, branch-polarity errors, Boolean-structure corruption, and output-contract violations. The results show that point accuracy alone is insufficient for characterizing LLM reliability in assertion generation and motivate robustness-aware evaluation for AI-assisted hardware verification.
\end{abstract}

\noindent\textbf{Keywords:} SystemVerilog assertions, hardware verification, large language models, metamorphic testing, robustness, RTL.

\section{Introduction}

Assertion-based verification (ABV) is a widely used mechanism for expressing design intent as executable properties. In SystemVerilog, assertions are used in simulation and formal verification to check that a register-transfer level (RTL) design obeys temporal and logical requirements. The difficulty is not merely writing syntactically valid SVA. A useful assertion must encode the correct control-flow precondition, timing relation, and consequent for the behavior being checked.

Recent work has increasingly applied machine learning and LLMs to this task. Earlier systems translated natural-language requirements into assertions using combinations of rules and learned models \cite{aditi2022hybrid,aditi2023validatable}. Subsequent LLM-based approaches have targeted security assertions, complete design specifications, structured specification/RTL representations, and domain-specific datasets \cite{kande2024security,yan2025assertllm,bai2025assertionforge,menon2025vert}. These systems show that LLMs can produce useful hardware-verification artifacts, but the predominant evaluation pattern remains \emph{point correctness}: a model receives one representation of an input and the generated assertion is judged correct or incorrect.

Point correctness does not answer a second, practically important question: \emph{is a correct generation stable under a semantics-preserving rewrite of the input?} Consider a condition such as
\begin{lstlisting}[language={}]
if (a && b && c) begin
    x = y;
end
\end{lstlisting}
Reordering the homogeneous conjunction to \texttt{c \&\& b \&\& a} does not change the Boolean condition. Similarly, consistently renaming identifiers or adding redundant parentheses should not alter the path condition that an SVA must capture. If the model changes a correct assertion into an incorrect one after such a rewrite, the original success is representation-sensitive.

This distinction matters in hardware verification. RTL is routinely reformatted, refactored, generated, renamed, or normalized by tools and engineers. Two source fragments may represent the same design behavior while differing substantially at the token level. A verification assistant that succeeds only for one surface form can appear accurate on a benchmark while remaining brittle in deployment.

Metamorphic testing provides a natural lens for this problem. Rather than asking only whether a single output is correct, it examines necessary relations between outputs on related inputs \cite{chen2018metamorphic}. For a semantics-preserving RTL transformation, the desired relation is straightforward: the generated assertion behavior should remain semantically correct. Recent literature has applied metamorphic testing to deep code models using transformations such as identifier changes and structural rewrites \cite{asgari2025metamorphic}, but this robustness perspective has received little attention in LLM-based SVA generation.

This paper therefore studies the following question: \emph{to what extent is LLM-generated SVA correctness invariant to semantics-preserving RTL representations?} We perform a controlled experiment over a stratified subset of VERT \cite{menon2025vert} using two open code models, an identical evaluation prompt, deterministic decoding, three input transformations, a behavior-level semantic scorer, and clustered bootstrap statistics.

The contributions are:
\begin{itemize}[leftmargin=*,nosep]
    \item a metamorphic evaluation framework for measuring representation sensitivity in RTL-to-SVA generation;
    \item three controlled semantics-preserving RTL transformations covering operand order, identifier names, and redundant parenthesization;
    \item robustness metrics that distinguish aggregate accuracy from retained correctness, correct-to-wrong failures, and total prediction flips; and
    \item an empirical study showing 9.7\%--27.0\% invariance failure across six model--transformation conditions, including cases where aggregate accuracy improves while previously correct behaviors regress.
\end{itemize}

The central claim is deliberately narrow. We do not infer that a model ``does not understand'' RTL semantics, nor do we attempt to identify the cause of a particular failure. We show that, on the controlled set studied here, SVA correctness is materially sensitive to source-level representations that preserve the intended Boolean behavior.

\section{Background and Related Work}

\subsection{SVA Generation}

For the conditional-control patterns considered here, assertion generation requires reconstructing the path condition under which an assignment is executed. For example:
\begin{lstlisting}[language={}]
if (a) begin
    x = y;
end else if (b) begin
    x = z;
end
\end{lstlisting}
The second assignment is guarded not merely by \texttt{b}, but by \texttt{!a \&\& b}. A model that omits the negation of an earlier branch produces a logically weaker and therefore incorrect assertion for that assignment behavior.

Aditi and Hsiao explored natural-language-to-SVA generation using hybrid rule-based and machine-learning techniques \cite{aditi2022hybrid} and later a validatable generation pipeline \cite{aditi2023validatable}. Kande et al. evaluated LLMs for security-focused hardware assertion generation and built a large automated evaluation framework \cite{kande2024security}. AssertLLM processes complete design specifications using multiple LLM-driven stages \cite{yan2025assertllm}. More recent work has incorporated RTL structure, knowledge graphs, progressive regularization, or richer evaluation signals \cite{bai2025assertionforge,wu2025spec2assertion}.

VERT directly targets SystemVerilog assertion generation by providing a large open dataset of RTL/SVA pairs and evaluating fine-tuned open models \cite{menon2025vert}. VERT is particularly useful for this study because it includes diverse conditional structures, synchronous and asynchronous variants, and explicit assertion references. Our work does not propose another generation architecture or fine-tuning method. Instead, it uses VERT as the substrate for a robustness study: given a model that produces a particular level of correctness on the original RTL, how much of that correctness survives semantics-preserving input rewrites?

The broader EDA literature increasingly treats LLMs as tools for code generation, verification, debugging, and knowledge retrieval \cite{he2025eda}. This breadth makes reliability evaluation increasingly important: a model output may be syntactically plausible and benchmark-correct while still being unstable under innocuous representation changes.

\subsection{Metamorphic Testing and Code-Model Robustness}

Metamorphic testing was introduced to address settings in which individual outputs are difficult to judge directly by instead checking necessary relations between multiple executions \cite{chen2018metamorphic}. The technique has since been applied across many software and machine-learning domains. In the context of source code, semantics-preserving transformations are especially attractive because they permit controlled perturbations while retaining program behavior.

A recent systematic review of metamorphic testing for deep code models identifies variable renaming and other semantics-preserving code transformations as recurring mechanisms for evaluating robustness \cite{asgari2025metamorphic}. The intuition carries naturally to hardware-description languages: identifier names, redundant grouping syntax, and the order of commutative Boolean operands can change the token sequence without changing the intended behavior.

The present study differs from standard code-generation metamorphic testing in two ways. First, the object being generated is a formal property whose antecedent has an explicit logical relationship to RTL control flow. Second, correctness can be decomposed at the assignment-behavior level, allowing us to distinguish aggregate gains from losses of behaviors that were already correct. This makes correct-to-wrong transitions a first-class metric rather than merely a change in top-line accuracy.

\section{Study Design}

\subsection{Research Questions}

We organize the study around three research questions:

\noindent\textbf{RQ1: Aggregate accuracy.} How does overall SVA behavior-level accuracy change under semantics-preserving RTL transformations?

\noindent\textbf{RQ2: Invariance.} Among behaviors that are correct on the original RTL, what fraction remain correct after transformation?

\noindent\textbf{RQ3: Failure modes.} What qualitative errors appear in sampled correct-to-wrong transitions?

\begin{figure*}[t]
\centering
\resizebox{0.98\textwidth}{!}{%
\begin{tikzpicture}[
  node distance=1.0cm,
  box/.style={draw, rounded corners=3pt, align=center, minimum height=2.7cm, text width=3.55cm, inner sep=8pt},
  arrow/.style={-{Latex[length=3mm]}, thick}
]
\node[box] (set) {\textbf{Dataset and evaluation set}\\[3pt]
20,000 VERT records\\10,400 conditional candidates\\9,157 eligible records\\40 controlled RTL programs};

\node[box, right=of set] (trans) {\textbf{Semantics-preserving transforms}\\[3pt]
T1: operand reordering\\
T2: identifier renaming\\
T3: redundant parentheses};

\node[box, right=of trans] (model) {\textbf{Controlled generation}\\[3pt]
\qwen{}\\
\deepseek{}\\[3pt]
Identical prompt; greedy decoding};

\node[box, right=of model] (score) {\textbf{Scoring and analysis}\\[3pt]
Behavior-level semantic scoring\\
Accuracy, CR, IF, flips\\
Clustered bootstrap CIs};

\draw[arrow] (set) -- (trans);
\draw[arrow] (trans) -- (model);
\draw[arrow] (model) -- (score);
\end{tikzpicture}%
}
\caption{Controlled metamorphic evaluation. VERT records are filtered to the supported conditional-control fragment before constructing the stratified 40-program evaluation set. Each applicable original/transformed pair is then processed with the same model, prompt, and generation settings. The transformation changes source representation while preserving the intended Boolean behavior.}
\label{fig:pipeline}
\end{figure*}
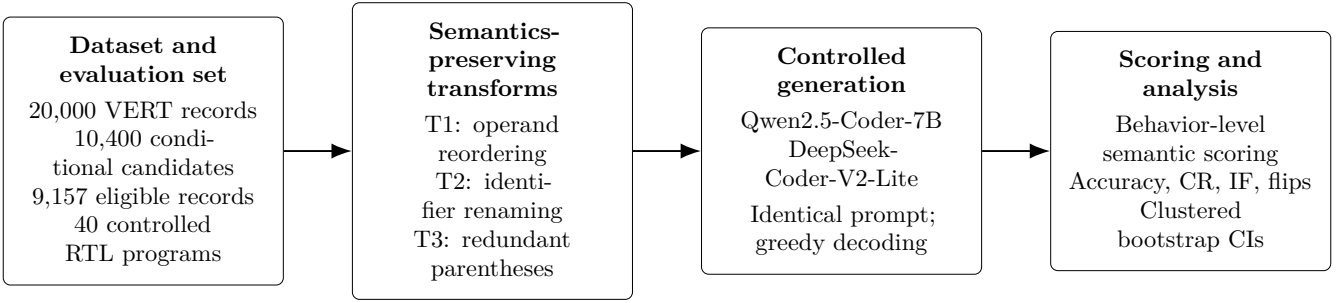

Figure~\ref{fig:pipeline} summarizes the controlled pipeline.

\subsection{Dataset Preprocessing and Quality Filtering}

We use the publicly released VERT dataset \cite{menon2025vert}, which contains 20,000 RTL/SVA records in the version audited for this study, evenly split between 10,000 synchronous and 10,000 asynchronous examples. Each record provides an RTL code fragment, one or more reference assertions, a synchronous/asynchronous indicator, and clock information when applicable.

We focus on conditional-control structures represented by nested \texttt{if} statements and multi-branch \texttt{if/else} trees because these programs expose the reasoning problem studied here: reconstructing the complete Boolean path condition under which an assignment executes. Exploratory auditing of case-style families revealed frequent dependence on X/Z wildcard semantics together with benchmark-quality issues such as malformed property syntax, missing semicolons, and unbalanced parentheses. Those families are therefore outside the controlled Boolean scope of this study rather than being treated as part of the quality-filtered conditional pool.

Restricting VERT to the nested-\texttt{if} and \texttt{if/else}-tree families produces 10,400 conditional-control candidates. Within this pool, we apply a quality filter that removes records with duplicate property names and/or an unexpected property count, because either condition prevents reliable one-to-one decomposition into the assignment behaviors required by the evaluator. This excludes 1,243 records and leaves 9,157 eligible examples, corresponding to 88.05\% of the conditional-control candidate pool.

The filtering procedure is intended to define a controlled fragment for which the reference behaviors, transformations, and semantic scorer can be applied consistently. It should not be interpreted as a claim that records outside this fragment are unsuitable for SVA generation in general.

\subsection{Controlled Evaluation Set}

The 9,157 eligible records form the \emph{source pool}; they are not all evaluated in the controlled experiment. To keep inference cost manageable while reducing dominance by repeated or near-identical templates, we construct a 40-program evaluation set using unique normalized structural templates. The set is stratified equally across four groups: nested-\texttt{if} asynchronous, nested-\texttt{if} synchronous, \texttt{if/else}-tree asynchronous, and \texttt{if/else}-tree synchronous. Each stratum contributes 10 programs.

Across these 40 RTL programs, the reference assertions decompose into 295 individual assignment behaviors. T1 is applicable to 38 programs and 280 behaviors because two programs contain no eligible homogeneous top-level conjunction or disjunction to reorder. T2 and T3 apply to all 40 programs and all 295 behaviors. Table~\ref{tab:dataset} summarizes the complete preprocessing and sampling progression.

\begin{table}[t]
\centering
\caption{Dataset construction and experimental scope.}
\label{tab:dataset}
\small
\begin{tabular}{lr}
\toprule
\textbf{Stage} & \textbf{Count} \\
\midrule
Raw VERT records & 20,000 \\
Synchronous / asynchronous & 10,000 / 10,000 \\
Conditional-control candidates & 10,400 \\
Excluded by quality filtering & 1,243 \\
Quality-filtered eligible pool & 9,157 \\
Controlled RTL programs & 40 \\
Assignment behaviors (T2/T3) & 295 \\
T1-applicable programs & 38 \\
T1 assignment behaviors & 280 \\
\bottomrule
\end{tabular}
\end{table}

The resulting evaluation set is intentionally a controlled sample rather than a population estimate over all VERT records. Statistical resampling therefore treats the RTL program/pair as the experimental unit rather than treating the hundreds of assignment behaviors as independent observations.

\subsection{Semantics-Preserving Transformations}

Table~\ref{tab:transforms} defines the three transformations.

\begin{table*}[t]
\centering
\caption{Controlled RTL transformations. The examples are schematic; the implementation transforms the full RTL condition while preserving its intended Boolean meaning.}
\label{tab:transforms}
\small
\begin{tabularx}{\textwidth}{p{0.08\textwidth}p{0.19\textwidth}X X}
\toprule
\textbf{ID} & \textbf{Transformation} & \textbf{Original example} & \textbf{Transformed example} \\
\midrule
T1 & Operand reordering & \texttt{a \&\& b \&\& c} & \texttt{c \&\& b \&\& a} \\
T2 & Identifier renaming & \texttt{valid \&\& ready} & \texttt{id\_000 \&\& id\_001} \\
T3 & Redundant parentheses & \texttt{a || b \&\& c} & \texttt{((a) || (b) \&\& (c))} \\
\bottomrule
\end{tabularx}
\end{table*}

\paragraph{T1: operand reordering.}
We reverse the operands of safe, homogeneous top-level logical conjunctions or disjunctions. The transformation does not change operators or introduce distributive rewrites. Restricting T1 to homogeneous commutative groups avoids ambiguity about operator precedence and makes the metamorphic relation explicit: the Boolean condition is unchanged even though local token order changes substantially.

\paragraph{T2: deterministic identifier renaming.}
Identifiers are alpha-renamed deterministically to neutral names such as \texttt{id\_000}. The mapping is consistent within a program. Because generated assertions must ultimately be compared with the original reference behaviors, model outputs are inverse-mapped before scoring. This separates sensitivity to identifier surface forms from changes in the underlying control/data relationships.

\paragraph{T3: redundant parenthesization.}
We add syntactically redundant parentheses around conditions and operands while preserving the non-parenthesis token sequence and operator order. The goal is not to alter precedence, but to test whether additional grouping syntax changes the model's interpretation of an otherwise identical logical expression.

These transformations are deliberately simple. Their value is that a robustness failure is easy to interpret: the source representation changed, but the target assignment behavior and its intended path condition did not.

\subsection{Models and Evaluation Prompt}

We evaluate two open code-oriented instruction models:
\begin{itemize}[leftmargin=*,nosep]
    \item \path{mlx-community/Qwen2.5-Coder-7B-Instruct-4bit}, based on \qwen{} \cite{hui2024qwen}; and
    \item \path{mlx-community/DeepSeek-Coder-V2-Lite-Instruct-4bit}, based on the 16B-total / 2.4B-active Mixture-of-Experts \deepseek{} model \cite{zhu2024deepseek}.
\end{itemize}

Both models are run locally with MLX on the same Apple M1 system with 16~GB memory. We use greedy decoding (temperature 0) and a maximum generation budget of 1,536 tokens. The identical evaluation prompt is used for both models and for every original and transformed RTL input.

The prompt requires one property per assignment behavior and instructs the model to derive a concrete Boolean antecedent from the complete RTL control-flow path. It requires enclosing branch conditions, negation of earlier branches when needed for \texttt{else-if} paths, exactly one implication operator per property, equality in the consequent, preservation of Boolean precedence, and property-only output. Timing is determined from the dataset record: synchronous examples use the supplied clock event and \texttt{|->}, whereas asynchronous examples use no event control and \texttt{|=>}. This convention is consistent across the 9,157-example eligible pool. Appendix~\ref{app:prompt} summarizes the operative prompt specification.

All numerical results reported in this paper use this identical prompt and the same deterministic generation configuration across original and transformed inputs.

\subsection{Behavior-Level Semantic Scoring}

We score each assignment behavior independently. A generated property must satisfy the output contract and match the target assignment's left- and right-hand sides. Its antecedent is then parsed into a restricted Boolean grammar containing negation, conjunction, disjunction, parentheses, and equality/inequality comparisons. The scorer checks propositional equivalence by exhaustive truth-table evaluation when the expression contains at most 16 Boolean atoms.

The scorer canonicalizes logically equivalent comparison forms. In particular, equality and inequality are normalized so that \texttt{a == b} can be compared with \texttt{!(a != b)}. Under the study's Boolean abstraction, identifier comparisons with 0/1 are also normalized, e.g., \texttt{!a} and \texttt{a == 0}. This abstraction is intentionally narrower than full SystemVerilog four-state semantics and is discussed as a limitation in Section~\ref{sec:limitations}.

For a behavior to be marked correct, the generated property must therefore satisfy three conditions: (1) it obeys the task's property/timing contract, (2) its consequent matches the assignment behavior, and (3) its antecedent is logically equivalent to the reference path condition under the scorer's Boolean abstraction. Combining multiple assignment behaviors into a single consequent violates the specified one-property-per-behavior contract even if the conjunction of consequents is logically compatible with a grouped reference.

\subsection{Metrics}

Let $N$ be the number of evaluated assignment behaviors for a transformation, $B$ the number correct on the original RTL, $T$ the number correct after transformation, $R$ the number correct in both conditions, $L$ the number that transition from correct to wrong, and $G$ the number that transition from wrong to correct. Thus $B=R+L$ and $T=R+G$.

We report baseline and transformed accuracy,
\begin{equation}
A_B = \frac{B}{N}, \qquad A_T = \frac{T}{N},
\end{equation}
and the accuracy delta $\Delta A=A_T-A_B$.

To measure stability of previously correct behaviors, we define conditional robustness
\begin{equation}
\crrate = \frac{R}{B},
\end{equation}
and invariance failure
\begin{equation}
\ifrate = \frac{L}{B}=1-\crrate.
\end{equation}
Finally, any-flip rate measures all correctness-state changes:
\begin{equation}
\fliprate = \frac{L+G}{N}.
\end{equation}

The distinction between $\Delta A$ and $\ifrate$ is central. A transformation can produce more gains than losses, yielding a positive aggregate accuracy delta, while still breaking a substantial fraction of behaviors that were originally correct.

\subsection{Statistical Analysis}

Assignment behaviors within one RTL program are correlated because they share the same source program and model generation. We therefore avoid treating all 280--295 behaviors as independent experimental units. Confidence intervals are computed with a clustered nonparametric bootstrap that resamples RTL pairs with replacement. We use 10,000 bootstrap replicates. T1 uses 38 clusters; T2 and T3 use 40.

We report percentile 95\% confidence intervals for accuracy delta, invariance failure, and any-flip rate. These intervals are used descriptively; in particular, a bootstrap interval for invariance failure that lies above zero is not presented as a formal null-hypothesis significance test.

\section{Results}

\subsection{RQ1: Aggregate Accuracy}

Table~\ref{tab:mainresults} presents the primary controlled results. \qwen{} begins at approximately 70\% behavior-level accuracy on the original RTL. T1 reduces accuracy from 70.0\% to 67.1\% ($-2.9$ percentage points), T2 from 69.8\% to 68.8\% ($-1.0$ point), and T3 from 69.8\% to 58.6\% ($-11.2$ points).

\deepseek{} begins lower, at 53.2--53.9\% baseline accuracy, but the direction of aggregate change differs by transformation. T1 increases accuracy by 5.0 points, T2 increases it by 9.8 points, and T3 decreases it by 4.1 points. Thus, the simple hypothesis that semantics-preserving rewrites always reduce aggregate accuracy is not supported.

\begin{table*}[t]
\centering
\caption{Controlled evaluation results. ``Lost'' denotes baseline-correct behaviors that become wrong after transformation; ``Gained'' denotes the reverse. $\Delta$ is transformed minus baseline accuracy in percentage points. CR = conditional robustness; IF = invariance failure.}
\label{tab:mainresults}
\scriptsize
\resizebox{\textwidth}{!}{%
\begin{tabular}{llrrrrrrrrrr}
\toprule
\textbf{Model} & \textbf{Transform} & \textbf{Pairs} & \textbf{Beh.} & \textbf{Base.} & \textbf{Trans.} & \textbf{$\Delta$} & \textbf{Lost} & \textbf{Gained} & \textbf{CR} & \textbf{IF} & \textbf{Flip} \\
\midrule
\qwen{} & T1 operand order & 38 & 280 & 70.0\% & 67.1\% & $-2.9$ & 19 & 11 & 90.3\% & 9.7\% & 10.7\% \\
\qwen{} & T2 identifier rename & 40 & 295 & 69.8\% & 68.8\% & $-1.0$ & 21 & 18 & 89.8\% & 10.2\% & 13.2\% \\
\qwen{} & T3 parentheses & 40 & 295 & 69.8\% & 58.6\% & $-11.2$ & 45 & 12 & 78.2\% & 21.8\% & 19.3\% \\
\midrule
\deepseek{} & T1 operand order & 38 & 280 & 53.2\% & 58.2\% & $+5.0$ & 24 & 38 & 83.9\% & 16.1\% & 22.1\% \\
\deepseek{} & T2 identifier rename & 40 & 295 & 53.9\% & 63.7\% & $+9.8$ & 31 & 60 & 80.5\% & 19.5\% & 30.8\% \\
\deepseek{} & T3 parentheses & 40 & 295 & 53.9\% & 49.8\% & $-4.1$ & 43 & 31 & 73.0\% & 27.0\% & 25.1\% \\
\bottomrule
\end{tabular}%
}
\end{table*}

The clustered bootstrap intervals in Table~\ref{tab:bootstrap} reinforce this caution. Only \qwen{} under T3 has an accuracy-delta interval excluding zero: $-11.2$ points with a 95\% interval of $[-21.3,-2.2]$. The other five intervals include zero. With only 38--40 program-level clusters, the aggregate deltas are imprecisely estimated and should not be over-interpreted.

\begin{table*}[t]
\centering
\caption{Clustered bootstrap statistics (10,000 replicates; resampling unit = RTL pair). Values are point estimate [95\% percentile CI]. Accuracy delta is in percentage points.}
\label{tab:bootstrap}
\small
\begin{tabular}{llccc}
\toprule
\textbf{Model} & \textbf{Transform} & \textbf{Accuracy $\Delta$} & \textbf{Invariance failure} & \textbf{Any-flip rate} \\
\midrule
\qwen{} & T1 & $-2.9$ [$-10.4$, $+2.9$] & 9.7\% [2.6, 19.3] & 10.7\% [4.7, 18.4] \\
\qwen{} & T2 & $-1.0$ [$-7.3$, $+5.2$] & 10.2\% [4.7, 16.6] & 13.2\% [8.3, 18.5] \\
\qwen{} & T3 & $-11.2$ [$-21.3$, $-2.2$] & 21.8\% [10.5, 34.1] & 19.3\% [10.7, 28.9] \\
\midrule
\deepseek{} & T1 & $+5.0$ [$-5.2$, $+15.6$] & 16.1\% [8.4, 24.5] & 22.1\% [14.9, 30.3] \\
\deepseek{} & T2 & $+9.8$ [$-2.5$, $+21.6$] & 19.5\% [8.8, 30.6] & 30.8\% [23.0, 38.9] \\
\deepseek{} & T3 & $-4.1$ [$-13.7$, $+6.1$] & 27.0\% [17.1, 36.9] & 25.1\% [17.7, 32.8] \\
\bottomrule
\end{tabular}
\end{table*}

\subsection{RQ2: Correctness Is Not Invariant}

Aggregate accuracy obscures the strongest pattern in the experiment. In every model--transformation condition, some behaviors that are correct on the original RTL become incorrect after transformation. Invariance failure ranges from 9.7\% to 27.0\%.

For \qwen{}, T1 and T2 preserve roughly 90\% of baseline-correct behaviors, corresponding to invariance-failure rates of 9.7\% and 10.2\%. T3 is substantially more disruptive: 45 of 206 baseline-correct behaviors become wrong, for an invariance-failure rate of 21.8\%.

\deepseek{} is more representation-sensitive on this controlled set. Invariance failure is 16.1\% for T1, 19.5\% for T2, and 27.0\% for T3. The largest any-flip rate occurs under T2: 30.8\% of all evaluated behaviors change correctness state.

Figure~\ref{fig:ifci} shows invariance-failure estimates with clustered bootstrap intervals. The intervals are relatively wide because the experimental unit is the program-level pair, but every condition exhibits a nontrivial loss of previously correct behavior.

\begin{figure*}[t]
\centering
\begin{tikzpicture}
\begin{axis}[
    width=0.93\textwidth,
    height=6.2cm,
    ymin=0, ymax=42,
    ylabel={Invariance failure (\%)},
    xlabel={Model--transformation condition},
    xtick={1,2,3,4,5,6},
    xticklabels={Qwen T1,Qwen T2,Qwen T3,DeepSeek T1,DeepSeek T2,DeepSeek T3},
    x tick label style={rotate=20,anchor=east},
    ymajorgrids=true,
    grid style={dashed,gray!30},
    error bars/error bar style={thick},
    error bars/error mark options={rotate=90,mark size=3pt},
    mark size=2.7pt,
]
\addplot+[
    only marks,
    mark=*,
    thick,
    error bars/.cd,
    y dir=both,
    y explicit,
] table[x=x,y=y,y error plus=ep,y error minus=em] {ifci.dat};
\end{axis}
\end{tikzpicture}
\caption{Invariance failure: fraction of baseline-correct assignment behaviors that become incorrect after a semantics-preserving transformation. Error bars are 95\% clustered bootstrap percentile intervals.}
\label{fig:ifci}
\end{figure*}
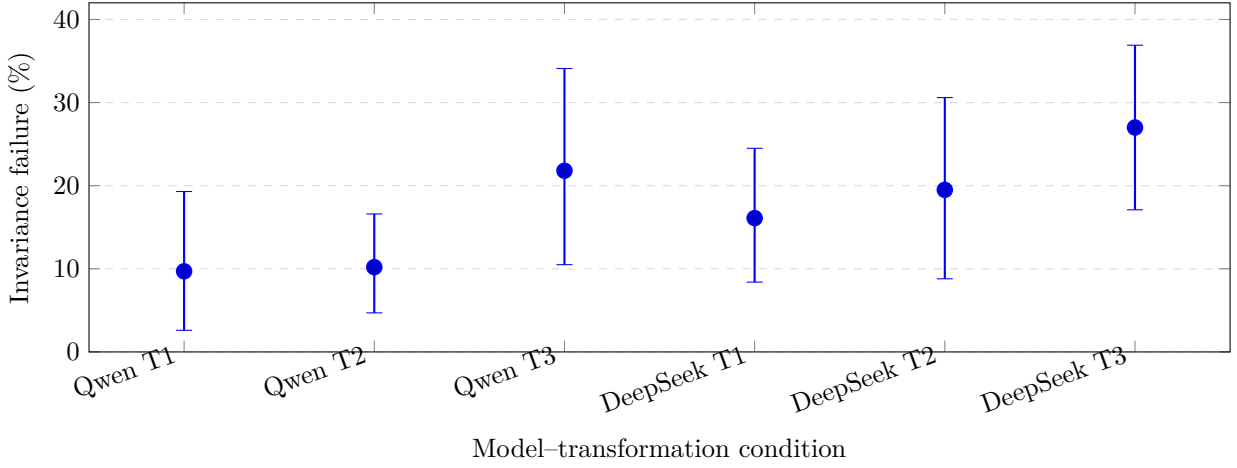

The DeepSeek T2 result illustrates why this metric is needed. Baseline accuracy is 53.9\% (159/295), and transformed accuracy rises to 63.7\% (188/295). The transformation creates 60 wrong-to-correct gains but also 31 correct-to-wrong losses. The net effect is positive aggregate accuracy, yet nearly one fifth of previously correct behaviors fail. A benchmark reporting only 53.9\% versus 63.7\% would therefore characterize the transformation as beneficial while missing substantial instability.

Figure~\ref{fig:deltaif} visualizes this decoupling. Conditions to the right of the vertical zero line improve aggregate accuracy, but their invariance failure remains high. Positive $\Delta A$ is therefore not evidence of representation invariance.

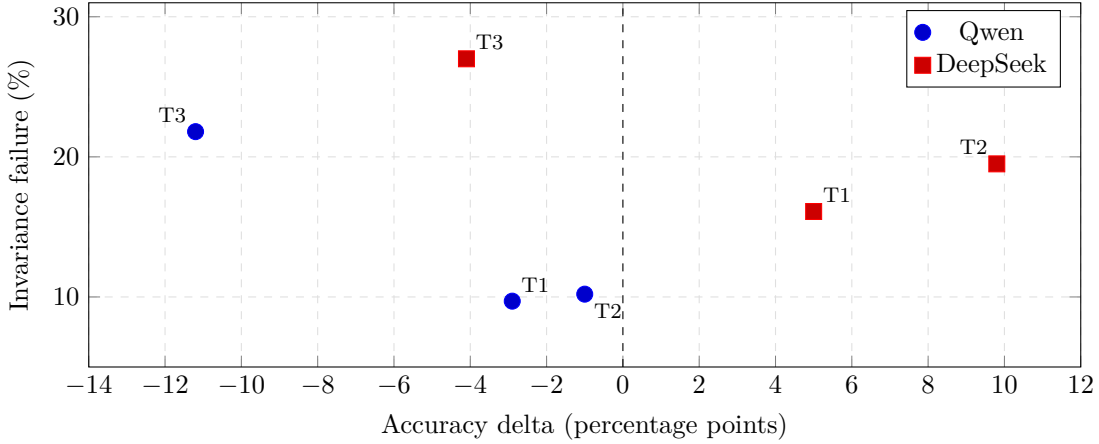
\begin{figure*}[t]
\centering
\begin{tikzpicture}
\begin{axis}[
    width=0.82\textwidth,
    height=6.4cm,
    xmin=-14, xmax=12,
    ymin=5, ymax=31,
    xlabel={Accuracy delta (percentage points)},
    ylabel={Invariance failure (\%)},
    xmajorgrids=true,
    ymajorgrids=true,
    grid style={dashed,gray!25},
]
\addplot+[only marks,mark=*,mark size=3pt] coordinates {
 (-2.9,9.7)
 (-1.0,10.2)
 (-11.2,21.8)
};
\addlegendentry{Qwen}
\addplot+[only marks,mark=square*,mark size=3pt] coordinates {
 (5.0,16.1)
 (9.8,19.5)
 (-4.1,27.0)
};
\addlegendentry{DeepSeek}
\draw[dashed] (axis cs:0,5) -- (axis cs:0,31);
\node[anchor=south west,font=\footnotesize] at (axis cs:-2.9,9.7) {T1};
\node[anchor=north west,font=\footnotesize] at (axis cs:-1.0,10.2) {T2};
\node[anchor=south east,font=\footnotesize] at (axis cs:-11.2,21.8) {T3};
\node[anchor=south west,font=\footnotesize] at (axis cs:5.0,16.1) {T1};
\node[anchor=south east,font=\footnotesize] at (axis cs:9.8,19.5) {T2};
\node[anchor=south west,font=\footnotesize] at (axis cs:-4.1,27.0) {T3};
\end{axis}
\end{tikzpicture}
\caption{Aggregate accuracy change and invariance failure measure different properties. In particular, DeepSeek T1 and T2 improve aggregate accuracy while still losing 16.1\% and 19.5\% of baseline-correct behaviors, respectively.}
\label{fig:deltaif}
\end{figure*}

\subsection{RQ3: Failure Analysis}

To validate that the measured losses correspond to substantive generation failures rather than scorer artifacts, we manually inspect five correct-to-wrong transitions per transformation per model. This yields 30 inspected losses in total.

For \qwen{}, all 15 sampled losses are genuine generation failures under the evaluation contract. For \deepseek{}, 13 of 15 are semantic/path-condition failures and two are output-contract violations in which the transformed response combines multiple assignment behaviors into one consequent rather than emitting one property per behavior. Under the evaluation contract these are correctly counted as failures, but we distinguish them from logical antecedent errors. No sampled transition is judged to be a scorer false positive in the final audit.

Table~\ref{tab:failuremodes} summarizes recurring failure patterns.

\begin{table*}[t]
\centering
\caption{Representative failure modes observed in the 30 manually inspected correct-to-wrong transitions. Examples are summarized at the logical level rather than reproduced verbatim.}
\label{tab:failuremodes}
\small
\begin{tabularx}{\textwidth}{p{0.13\textwidth}p{0.09\textwidth}p{0.24\textwidth}X}
\toprule
\textbf{Model} & \textbf{Transform} & \textbf{Failure mode} & \textbf{Observed effect} \\
\midrule
Qwen & T1 & Dropped prior-branch guard & An \texttt{else-if} assignment loses the required negation of an earlier condition after operands are reordered. \\
Qwen & T2 & Wrong branch polarity & A renamed \texttt{else-if} path is generated using the previous disjunctive branch positively rather than requiring that branch to be false. \\
Qwen & T3 & Boolean-structure corruption & A negated disjunction is rewritten as a disjunction of negations, or an original OR relationship becomes an AND. \\
DeepSeek & T1 & Missing path predicate & A reordered condition causes one of the earlier branch constraints to disappear from the assignment antecedent. \\
DeepSeek & T2 & Output-contract violation & Multiple assignment behaviors are merged into a single conjunctive consequent instead of one property per behavior. \\
DeepSeek & T3 & Guard inversion / omission & Redundant parentheses trigger positive conditions to be negated, required guards to disappear, or prior-branch structure to be inverted. \\
\bottomrule
\end{tabularx}
\end{table*}

Three qualitative observations are notable. First, failures are not limited to exotic syntax. T1 can change only operand order while inducing missing guards. Second, T3 is particularly revealing because additional parentheses should, if anything, make Boolean grouping more explicit; nevertheless, both models show their highest invariance-failure rate under T3. Third, several errors involve control-flow composition rather than local syntax---for example, correctly negating all earlier branches before an \texttt{else-if} assignment. This suggests that robustness evaluation should preserve the full nested control context rather than test isolated Boolean expressions only.

\section{Analysis and Implications}

\subsection{Point Accuracy and Robustness Measure Different Properties}

The primary result is not that every transformation decreases accuracy; the data explicitly contradict that statement. Rather, aggregate accuracy and behavioral stability are different axes of model quality. A model can gain more behaviors than it loses and therefore improve overall accuracy while still regressing on inputs it previously handled correctly.

This distinction is important for comparing assertion-generation systems. Suppose model A scores 65\% on a fixed benchmark and model B scores 70\%. Without a robustness test, it is unclear whether either score reflects stable reasoning across semantically equivalent representations. Metamorphic evaluation adds a second question: of the behaviors the model gets right, how many remain right under controlled rewrites? Conditional robustness and invariance failure provide a direct answer.

For deployment, one possible implication is to treat representation agreement as an additional confidence signal. A verification assistant could generate assertions from several semantics-preserving variants of the same RTL and flag behaviors whose outputs disagree. Such a strategy would not prove correctness, but it could expose fragile generations before they enter a verification flow. Evaluating that intervention is beyond the scope of this paper.

\subsection{Transformation-Specific Sensitivity}

T3 produces the largest invariance failure for both models: 21.8\% for \qwen{} and 27.0\% for \deepseek{}. This is surprising because the transformation does not reorder non-parenthesis tokens or rename variables; it only adds grouping syntax. The result suggests sensitivity to token-level form even when the logical parse intended by the transformation is unchanged. However, the present experiment cannot determine whether the cause is tokenization, learned code-style priors, prompt interaction, quantization, or another property of inference.

T2 shows a different pattern. Identifier renaming has a relatively modest aggregate effect on \qwen{} ($-1.0$ point) and a strongly positive point estimate on \deepseek{} ($+9.8$ points), yet the corresponding invariance-failure rates are 10.2\% and 19.5\%. Neutral names may remove misleading lexical cues in some cases while simultaneously disrupting behaviors that benefited from those cues. Again, this is a plausible interpretation rather than a causal conclusion.

T1 demonstrates that even commutative local rewrites can alter downstream control-flow reconstruction. This matters because the target SVA is not merely a translation of the edited subexpression; it must integrate that expression with enclosing and preceding branch predicates.

\subsection{Implications for Hardware-Verification Benchmarks}

Current assertion-generation evaluations often emphasize syntax correctness, functional correctness, coverage, or exact/semantic agreement on a fixed benchmark \cite{kande2024security,yan2025assertllm,menon2025vert}. These remain necessary metrics. Our results suggest adding a robustness layer rather than replacing them.

A benchmark can report at least three complementary quantities: (1) point accuracy on canonical inputs, (2) conditional robustness of baseline-correct behaviors under semantics-preserving variants, and (3) total flip rate. This decomposition reveals whether performance gains arise from stable improvement or from a large reshuffling of which behaviors happen to be correct.

Metamorphic variants can also be useful when a benchmark contains synthetic or templated code. A model may exploit consistent naming, formatting, or expression style without those cues being part of the actual verification problem. Controlled rewrites provide a relatively inexpensive probe of such dependence.

\section{Threats to Validity and Limitations}
\label{sec:limitations}

\paragraph{Controlled sample size.}
The experimental sample contains 40 RTL programs, with 38 applicable to T1. This is sufficient to expose repeated representation-sensitive failures but does not support precise population-level estimates over all 20,000 VERT records. The clustered bootstrap intervals are correspondingly wide. We therefore frame the work as a controlled robustness study rather than a definitive ranking of models or transformations.

\paragraph{Dataset scope.}
The study focuses on the conditional-control subset of VERT, specifically nested conditionals and multi-branch \texttt{if/else} structures. The results may not generalize to case statements, temporal sequences, arithmetic-heavy RTL, protocol-level properties, or large industrial modules. VERT itself contains synthetic augmentation and has different characteristics from hand-written production RTL \cite{menon2025vert}.

\paragraph{Two models.}
We evaluate one dense 7B code model and one MoE code model using local 4-bit MLX checkpoints. Two models are insufficient to draw conclusions about architecture families, parameter count, or quantization. The purpose of using two models is to test whether representation sensitivity appears across distinct open code models under the same controlled protocol.

\paragraph{Boolean semantic abstraction.}
The scorer reasons propositionally over a restricted Boolean grammar and canonicalized equality/inequality expressions. It is not a full SystemVerilog simulator or formal engine and does not model four-state X/Z semantics, arbitrary bit-vector arithmetic, or all temporal SVA constructs. We mitigate this threat by filtering the dataset to a controlled fragment and by manually inspecting sampled losses, but a stronger future study should validate generations with a full compiler/formal/simulation stack.

\paragraph{Transformation validity.}
The transformations are designed to preserve the relevant Boolean behavior, and T1 is restricted to safe homogeneous operator groups. T2 applies consistent alpha-renaming with inverse mapping, and T3 changes only parenthesization. Nevertheless, any transformation framework can contain implementation bugs. Public release of the exact transformed inputs and scripts is therefore important for reproducibility.

\paragraph{Behavior-level dependence.}
A program contributes multiple assignment behaviors, so behavior outcomes are not independent. We address this in uncertainty estimation by resampling at the RTL-pair level. The raw behavior counts in Table~\ref{tab:mainresults} should not be interpreted as 280--295 independent experimental samples.

\paragraph{No causal claim.}
Observed sensitivity does not establish why a model changes its output. We do not attribute failures to memorization, contamination, tokenization, architecture, or a lack of semantic understanding. The experiment demonstrates invariance failure under controlled rewrites; causal diagnosis is future work.

\section{Reproducibility}
\label{sec:repro}

Both models are evaluated using the same controlled dataset, transformation procedures, evaluation prompt, and deterministic generation configuration. All controlled generations use greedy decoding with temperature 0 and a maximum generation length of 1,536 tokens. Confidence intervals are computed with 10,000 clustered bootstrap resamples at the RTL-program level, and the manual failure analysis examines five correct-to-wrong transitions from each model--transformation condition.

To support independent reproduction, the accompanying experimental artifact is intended to include the dataset-preprocessing and transformation scripts, controlled evaluation set, transformed RTL inputs, exact evaluation prompt, raw model generations, paired original/transformed records, behavior-level scoring outputs, bootstrap analysis, and manual failure-analysis samples. These materials provide the implementation-level details omitted from the manuscript for readability.

\section{Conclusion}

This paper evaluates a property that conventional SVA-generation accuracy does not capture: whether correctness survives semantics-preserving changes in RTL representation. On a stratified 40-program conditional-control subset of VERT, two open code models exhibit correct-to-wrong failures under operand reordering, deterministic identifier renaming, and redundant parenthesization. Depending on the model and transformation, 9.7\%--27.0\% of baseline-correct assignment behaviors become incorrect.

The results also show why aggregate accuracy alone is insufficient. \deepseek{} improves by 9.8 percentage points under identifier renaming while losing 19.5\% of behaviors it originally answered correctly. Robustness must therefore be measured directly rather than inferred from a net accuracy change.

The study is intentionally controlled and limited in scale. Future work should extend the evaluation to larger and more diverse RTL corpora, additional model families, richer temporal SVA constructs, and full formal/simulation-based semantic validation. More broadly, metamorphic testing offers a practical way to ask not only whether an AI verification assistant is correct on a benchmark input, but whether that correctness is stable when irrelevant surface details change.

\appendix
\section{Evaluation Prompt Specification}
\label{app:prompt}

The same prompt specification is applied to both models and to all original and transformed RTL inputs. The operative requirements are summarized below; the exact prompt text used in the experiments is retained with the accompanying experimental artifact.

\begin{itemize}[leftmargin=*]
    \item Generate one SystemVerilog property for each assignment behavior represented in the RTL.
    \item Derive a concrete Boolean antecedent from the complete control-flow path; do not emit placeholders or generic condition names.
    \item Include enclosing branch conditions and, for \texttt{else-if} paths, negate all earlier mutually exclusive branches as required by the RTL control flow.
    \item Preserve Boolean operator precedence. When negating a compound expression, negate the complete expression rather than changing its internal Boolean structure.
    \item Use exactly one implication operator per property and express the consequent as an equality matching the target assignment behavior.
    \item For synchronous records, use the clock event supplied with the record and \texttt{|->}; for asynchronous records, emit no event control and use \texttt{|=>}.
    \item Emit property declarations only: no procedural \texttt{assert}, \texttt{if}, \texttt{begin}/\texttt{end}, explanatory prose, or Markdown formatting.
\end{itemize}

\section*{Acknowledgment of AI Assistance}
OpenAI ChatGPT was used to assist with manuscript organization, language drafting, and presentation. The experimental design, execution, numerical results, technical verification, interpretation, and final manuscript decisions remain the responsibility of the author. Bibliographic entries used in the manuscript were checked against public publication records before inclusion.

\bibliographystyle{unsrtnat}
\bibliography{references}

\end{document}